\documentclass[6pt]{article}
\usepackage{fancyvrb}
\usepackage[margin=.95in]{geometry}
\usepackage{xcolor}
\usepackage{multicol}
\usepackage[skip=.5em,position=bottom]{caption}

\newcommand{\myHrule}{\vspace{0.1em} \hrule \vspace{0.1em}}

\usepackage{inconsolata}

\usepackage[style=numeric]{biblatex}
\providecommand{\keywords}[1]{\textbf{Keywords:} #1}

\author{
  Laria Reynolds\\
  \texttt{moire@knc.ai}
  \and
  Kyle McDonell\\
  \texttt{kyle@knc.ai}
}

\date{}

\title{Prompt Programming for Large Language Models: \\
Beyond the Few-Shot Paradigm}

\begin{document}

\maketitle

\begin{abstract}
    Prevailing methods for mapping large generative language models to supervised tasks may fail to sufficiently probe models' novel capabilities. Using GPT-3 as a case study, we show that 0-shot prompts can significantly outperform few-shot prompts. We suggest that the function of few-shot examples in these cases is better described as locating an already learned task rather than meta-learning. This analysis motivates rethinking the role of prompts in controlling and evaluating powerful language models. In this work, we discuss methods of prompt programming, emphasizing the usefulness of considering prompts through the lens of natural language. We explore techniques for exploiting the capacity of narratives and cultural anchors to encode nuanced intentions and techniques for encouraging deconstruction of a problem into components before producing a verdict. Informed by this more encompassing theory of prompt programming, we also introduce the idea of a \emph{metaprompt} that seeds the model to generate its own natural language prompts for a range of tasks. Finally, we discuss how these more general methods of interacting with language models can be incorporated into existing and future benchmarks and practical applications.
\end{abstract}

\vspace{1em}

\begin{multicols}{2}

\keywords{language models, transformers, GPT-3, few-shot learning, prompt programming, metaprompts, serial reasoning, semiotics}

\section{Motivation}
 
The recent rise of massive self-supervised language models such as GPT-3 \cite{brown2020language} and their success on downstream tasks has brought us one step closer to the goal of task-agnostic artificial intelligence systems. However, despite the apparent power of such models, current methods of controlling them to perform specific tasks are extremely limited. In order to properly evaluate their capabilities and extract useful work from these models, new methods are required. \par 

Prior to GPT-3, the standard approach to the evaluation and use of such models has involved fine-tuning on a portion of a task dataset \cite{howard2018universal}. GPT-3 achieved state-of-the-art performance on a wide variety of tasks without fine tuning, using only \textit{few-shot} prompts, in which a small number of examples of solved tasks are provided as part of the input to the trained model. However, while the few-shot format was sufficient to reveal surprising performance on these tasks, we argue that prompting can be more effective than either fine-tuning or the few-shot format at extracting specific learned behaviors from self-supervised language models. \par

We argue that contrary to the common interpretation of the few-shot format implied by the title of the original GPT-3 paper \cite{brown2020language}, \textit{Language models are few-shot learners}, GPT-3 is often not actually \textit{learning} the task during run time from few-shot examples. Rather than instruction, the method's primary function is \textit{task location} in the model's existing space of learned tasks. This is evidenced by the effectiveness of alternative prompts which, with no examples or instruction, can elicit comparable or superior performance to the few-shot format. \par

This motivates new approaches which explicitly pursue the goal of task location. We propose exploring more general methods of prompt programming and specifically techniques for communicating task intention and structure to an self-supervised model in the modality it was trained: natural language.  \par

The ground truth function that self-supervised language models are trained to approximate is, in great generality, is how humans write. Accordingly, to interact with and control a language model, we should consider doing so from the perspective of natural language as it is used by humans. With a few caveats, we want to find prompts which we would expect a human to complete in a way that accomplishes the desired task.\par

In this paper, we investigate the few-shot paradigm and find that its performance can be matched or exceeded by simple 0-shot prompts. We explore the nature of successful 0-shot prompts and propose general methods of prompt programming through the lens of natural language semiotics. We demonstrate novel prompts which force a language model to break a problem into components before producing a verdict, and we introduce the concept of \textit{metaprompt programming}, an approach which offloads the job of writing a task-specific prompt to the language model itself. Finally, we discuss how these ideas can be incorporated into existing and future benchmarks to allow us to better probe the capabilities of large language models.

\section{Related work}

Recent work in the literature has focused on controlling natural language generation using traditional approaches from machine learning like novel architectures which condition outputs \cite{DBLP:journals/corr/abs-1909-05858, KrauseGeDi2020}, more advanced sampling techniques \cite{fan2018hierarchical, holtzman2020curious}, gradient-based optimization of prompts \cite{shin2020autoprompt, li2021prefix}, and task-specific adapter networks \cite{ye2021zeroshot}. See \cite{wang2021control} for a survey of these recent methods. Past work has also explored improving the few-shot paradigm by dynamically selecting the most relevant examples for each task \cite{liu2020multi, gao2020making}.\par

In comparison, little work on natural-language, 0-shot approaches to prompt programming has been formalized. Instead, successful prompt programming techniques have primarily been shared on blogs and social media among users of OpenAI's API and AI Dungeon. \par 

Due to the decentralized form that most explorations of prompt programming have taken, it is not feasible for us to to compile all relevant contributions here. We instead give a brief, non-exhaustive survey of explorations which have gone beyond the few-shot paradigm. \par

Gwern has given the most comprehensive survey of GPT-3's capabilities through demonstrations of it writing fiction, poetry, navy seals copypasta parodies, and performing tasks like PDF cleaning. He has written extensively about his intuitions of working with GPT-3 and his methods of prompt programming \cite{branwen2020gpt}. Arram Sabeti has written about the effect of the context provided by a prompt on writing quality \cite{sabetioutputquality}. Zachary Robertson has written about amplifying GPT-3's mathematical capabilities through a dialogue that guides it to break a problem into steps \cite{robertsonamplification}. Twitter user KaryoKleptid has posted experiments along a similar vein, using dialogues to prompt GPT-3 (via AI Dungeon) to break problems into steps and follow procedures such as brute force checking \cite{kleptidserial, kleptidbruteforce}, achieving impressive results on math problems.

Our work synthesizes and expands on the methods pioneering by these explorations, representing a modest step towards formalizing effective natural language prompt programming techniques. 

\section{Investigating \\few-shot prompting}\label{sec:few_shot}

GPT-3 was evaluated on tasks with 0, 1, and $n$-shot prompts (containing only a natural language description, one solved example, and $n$ solved examples respectively). GPT-3 consistently performs better when more examples are provided, with 0-shot performance often achieving less than half of the score of many-shot tests. A common interpretation of this result is that GPT-3 is learning from the examples at runtime and this allows it to perform better than with fewer or no examples \cite{brown2020language}. \par

The improvement in performance with the number of examples, however, can be interpreted in an alternate way. Rather than learning how to perform the task from the examples, the examples may simply serve to instruct GPT-3 on what task it is to solve and encourage it to follow the structure of the prompt in its response. \par

For example, for certain tasks, such as translation, a small number of samples is insufficient to learn anything substantial about the task. Instead, GPT-3 must rely primarily, if not entirely, on the knowledge of vocabulary and grammar of both the source and target languages embedded in its trained weights. Rather than viewing these tasks as \textit{few-shot-learning}, we will explicitly show that these prompts primarily direct the model to access existing knowledge. We do so by investigating whether examples (training samples) are even necessary.\par

\subsection{The success of 0-shot prompts}

Due to budget and time constraints, we explore a single illustrative example, a French-to-English translation task. We find that 0-shot prompts can match and even exceed standard few-shot performance. Our results in table \ref{tab:table} show that the 0-shot accuracy reported in the original GPT-3 paper \cite{brown2020language} can be improved substantially with even minor prompt engineering. Most significantly, the extremely simple prompt in Figure (\ref{fig:simple_colon}) which includes only the names of the two languages and a colon performs better than the 10-shot prompt in the style of the original GPT-3 paper. \par

In fact, we found this pattern was true of most of the worst-performing 0-shot prompts in the original GPT-3 paper \cite{brown2020language}, particularly question and answer benchmarks. Many could easily be improved by simple changes in formatting which make the prompt closer to natural language as a human would write it. Thus, GPT-3's 0-shot or baseline performance without meta-learning was significantly underestimated. \par

It is important to correct this confusion to get a more precise understanding of the nature of a model’s capabilities so that we can better learn to control it. The fact that GPT-3 has a vast repertoire of functions that do not need to be learned at runtime allows for great flexibility in 0-shot prompting and encourages exploring more general methods of prompt programming. \par

\subsection{Examples don't always help}\label{sec:examples_hurt}

In our experiment, the simple colon prompt (Figure \ref{fig:simple_colon}) 1-shot performed significantly worse than 0-shot. By examining the output of GPT-3 on this task we found that the decreased performance was due to semantic contamination from the 1-shot example. Instead of treating the example as a categorical guide, it is inferred that the semantic meaning of the examples are relevant to the task, e.g. the example is interpreted as part of a consecutive narrative. Indeed, we found this was true more generally of low-shot prompts across a variety of tasks. \par

This effect of contamination from few-shot examples has been successfully used to improve the performance of GPT-3 by selecting in-context examples for each task \cite{liu2020multi}. 

\end{multicols}

\vspace{.3em}

\begin{center}
    \begin{tabular}{|l|c|c|}
        \hline
        Prompt & Babbage / 6.7B & Curie / 13B \\ \hline \hline
        OpenAI 0-shot & 15.5 & 22.4 \\ \hline
        OpenAI 1-shot & 31.6 & 31.4 \\ \hline
        OpenAI 64-shot & 36.4 & 38.3 \\ \hline \hline
        Reproduced OpenAI 0-shot & 15.9 & 18.7  \\  \hline
        Reproduced OpenAI 1-shot & 21.8 & 24.1   \\ \hline
        Reproduced OpenAI 10-shot & 25.1 & 27.9  \\  \hline
        Simple colon 0-shot & 23.5 & 33.3   \\ \hline
        Simple colon 1-shot & 18.0 & 27.6  \\ \hline
        Simple colon 10-shot & 24.1 & 33.4   \\ \hline
        Master translator 0-shot & 26.5 & 32.9   \\ \hline
    \end{tabular}\\
    \vspace{0.5em}
    \captionof{table}{We report BLEU scores for variants of the GPT-3 model using different prompt formats on the WMT’14 Fr-En dataset \cite{bojar2014translation} as measured by SacreBLEU \cite{post2018clarity}. First are results reported in the original GPT-3 paper \cite{brown2020language} on the 6.7B and 13B parameter versions of GPT-3, our attempts to reproduce the results according to those exact specifications using the \textit{Babbage} and \textit{Curie} models available from OpenAI's API, and finally results from custom prompts described in (Figures \ref{fig:simple_colon},\ref{fig:master_translator}). The difference in the reproduced results may be attributable to changes in the OpenAI API after the publication of their results or because of unknown hyperparameters. Additionally, the size of the \textit{Babbage} and \textit{Curie} models are not reported so the relationship to the models in the original GPT-3 paper is inferred. We were unable to replicate the 64-shot test due to API constraints and instead replaced it with a 10-shot test. }
    \label{tab:table}
\end{center}


\begin{multicols}{2}
\begin{center}
    \footnotesize
    \centering
    \myHrule
\begin{Verbatim}[commandchars=\\\{\}]

  French: \textbf{example_source_phrase}
  English: \textbf{example_target_phrase}
   
  French: \textbf{example_source_phrase}
  English: \textbf{example_target_phrase}
    
  \textbf{[...]}
    
  French: \textbf{source_phrase}
  English: 
\end{Verbatim}
    \myHrule
    \captionof{figure}{The ``Simple Colon'' prompt format. For few-shot tasks, additional examples are provided as shown. Text in \textbf{bold} is to be replaced by source and target language text examples.}
    \label{fig:simple_colon}
\end{center}
\vspace{-1.0em}
\begin{center}
    \footnotesize
    \centering
    \myHrule
\begin{Verbatim}[commandchars=\\\{\}]

  A French phrase is provided: \textbf{source_phrase}
  The masterful French translator flawlessly 
  translates the phrase into English:
\end{Verbatim}
    \myHrule
    \captionof{figure}{The ``Master Translator'' prompt format. Text in \textbf{bold} is to be replaced by source and target language text examples.}
    \label{fig:master_translator}
\end{center}

\vfill\null
\columnbreak

\section{Prompt programming}

Rewriting a prompt can result in significant changes to the performance of a language model on tasks. That motivates the question: Is there a methodology which we can follow to craft prompts more likely to yield desired behavior?\par
Prompt engineering for a language model whose input and output are in natural language may be conceived as \textit{programming in natural language}. Natural language, however, is indeterministic and much more complex than traditional programming languages. In this section, we open a discussion about the theory and method of natural language programming.

\subsection{The dynamics of language}

To understand how to prompt an autoregressive language model, we must first consider the context in which it was trained and the function it approximates.\par

GPT-3 was trained in a self-supervised setting on hundreds of gigabytes of natural language \cite{brown2020language}. Self-supervision is a form of unsupervised learning in which ground truth labels are derived from the data itself. In the case of GPT-3, the ground truth label assigned to each example was simply the token that came next in the original source. The ground truth \textit{function} which GPT-3 approximates, then, is the underlying dynamic that determined what tokens came next in the original source. This function, unlike GPT-3, is not a black box - we live and think its components - but it is tremendously, intractably complex. It is the function of human language as it has been used and recorded by humans in books, articles, blogs, and internet comments. \par

A system which predicts the dynamics of language necessarily encompasses models of human behavior and the physical world \cite{lg2020agilms}. The ``dynamics of language'' do not float free of cultural, psychological, and physical context; it is not merely a theory of grammar or even of semantics. Language in this sense is not an abstraction but rather a phenomenon entangled with all aspects of human-relevant reality. The dynamic must predict how language is actually used, which includes (say) predicting a conversation between theoretical physicists. Modeling language is as difficult as modeling every aspect of reality that could influence the flow of language.\par 

GPT-3 has not learned the ground truth function perfectly, obviously, or else the world would look very different by now. However, it has approximated it to a notable extent, as evidenced by its ability to not only form grammatical sentences, but also coherently employ cultural references and metaphors and model complex psychological and physical contexts \cite{branwen2020gpt}. The problem of prompt programming, then, is nontrivial, for the dynamics of language (or an approximation thereof on GPT-3's level of sophistication) are nontrivial.\par 

If we were to predict how a given passage of text would continue given that a human had written it, we would need to model the intentions of its writer and incorporate worldly knowledge about its referents. The inverse problem of searching for a prompt that would produce a continuation or class of continuations involves the same considerations: like the art of persuasion, it entails high-level, mentalistic concepts like tone, implication, association, meme, style, plausibility, and ambiguity.\par

This motivates an anthropomorphic approach to prompt programming, since modelling how GPT-3 will react to a prompt involves modelling virtual human writer(s). An anthropomorphic approach is distinct from \textit{anthropomorphizing the model}. GPT-3's dynamics entail sophisticated predictions of humans, but it behaves unlike a human in several important ways. In this paper we will address two such ways: its resemblance not to a single human author but a superposition of authors, which motivates a subtractive approach to prompt programming (\S\ref{sec:constraining}), and its constrained ability to predict dynamics in situations where a substantial amount of silent reasoning happens between tokens, a limitation which can be partially overcome by prompting techniques (\S\ref{sec:serializing}).\par

The thrust of this section is that formulating an exact theory of prompt programming for a self-supervised language model belongs to the same difficulty class as writing down the Hamiltonian of the physics of observable reality (very hard). However, humans have an advantage to be effective at prompt programming nonetheless, because we have evolved and spent our lives learning heuristics relevant to the dynamics at hand. Prompt programming is programming in natural language, which avails us of an inexhaustible number of functions we know intimately but don’t have names for. We need to learn a new methodology, but conveniently, we've already learned the most difficult foundations. The art of prompt programming consists in adapting our existing knowledge to the peculiarities of interacting with an autoregressive language model.\par

In \S\ref{sec:direct} - \S\ref{sec:metaprompt}, we present methods and frameworks which we have found to be helpful for crafting effective prompts. These methods can and should be applied in parallel, just as they are woven together in all forms of human discourse. In general, the more redundancy reinforcing the desired behavior the better, as is arguably demonstrated by the effectiveness of the few-shot format. \par

As our experience derives primarily from interacting with GPT-3, in the following sections we refer directly and indirectly to the capabilities and behaviors of GPT-3. However, we believe that these methods generalize to prompting any autoregressive language model trained on a massive human-written corpus.

\subsection{Direct task specification: \\constructing the signifier}
\label{sec:direct}

Pre-GPT-3 models had much less capability to understand abstract descriptions of tasks due to their limited model of the world and human concepts. GPT-3's impressive performance on 0-shot prompts indicates a new realm of possibilities for direct task specification.\par

A direct task specification is a 0-shot prompt which tells the model to perform some task that it already knows how to do. A direct specification consists in constructing a \textit{signifier} for the task. A signifier is a pattern which keys the intended behavior. It could be the name of the task, such as ``translate'', a compound description, such as ``rephrase this paragraph so that a 2nd grader can understand it, emphasizing real-world applications'', or purely contextual, such as the simple colon prompt from Figure \ref{fig:simple_colon}. In none of these cases does the signifier explain \textit{how} to accomplish the task or provide examples of intended behavior; instead, it explicitly or implicitly calls functions which it assumes the language model has already learned.\par

Direct specifications can supervene on an infinity of implicit examples, like a closed-form expression on an infinite sequence, making them very powerful and compact. For instance, the phrase ``translate French to English'' supervenes on a list of mappings from all possible French phrases to English.\par

A large language model, like a person, has also learned behaviors for which it is less obvious how to construct a direct signifier. Task specification by demonstration (\S\ref{sec:demonstration}) and by proxy (\S\ref{sec:proxy}) may be viable alternative strategies for eliciting those behaviors.

\subsection{Task specification by \\demonstration}
\label{sec:demonstration}
Few-shot examples are effective for task specification because the pattern of sequential repetitions of a function with varying parameters is common to natural language. Unlike previous models, GPT-3 has learned this property of language robustly and is able to apply it in contrived situations when the examples are stripped of all context. Like direct specification, task specification by \textit{demonstration} is a possibility opened by GPT-3.

Some tasks are most effectively communicated using examples, such as when the task requires a bespoke format, the language in which the examples are described is better developed or understood than the meta-language required for a description of the task itself or very instructive examples are available.\par

It is important to note that unlike in fine-tuning, the ``training examples'' in few-shot are processed as a whole, and may not necessarily be interpreted as parallel and independent. Informative context or a large number of examples can help mitigate the problems with few-shot addressed in \S\ref{sec:examples_hurt}. For instance, a prompt could embed examples in a context which makes it clear that the examples are independent instances of a function rather than a sequential pattern that should be extrapolated. In general, examples are more efficient and informative in context, both from the perspective of a human and a language model \cite{aidungeonanalogy}.\par

\subsection{Task specification by memetic proxy}
\label{sec:proxy}

Another method used in human communication is proxies or analogies, where a memetic concept such as a character or characteristic situation is used as a proxy for an intention, the latter which may be quite complex or nuanced. GPT-3 demonstrates nuanced understanding of analogies \cite{aidungeonanalogy}. Specification by proxy is mechanistically similar to direct specification, except that the signifier keys behaviors from memespace/cultural consciousness instead of naming the behavior directly.\par

For instance, instead of specifying exact criteria for an answer to a moral question directly or using examples, you could ask Mahatma Gandhi, Ayn Rand, or Eliezer Yudkowksy. Each will come not only with a complex biases but also assumptions about the context of the question, which may be take paragraphs to demonstrate or describe. GPT-3’s ability to create simulations of well-known figures and to draw on cultural information far exceeds the ability of most humans \cite{branwen2020gpt}, so this method is particularly useful for encoding a complex (especially open-ended) task. Since GPT-3 lends itself well to embeddings in a narrative context, the infinite degrees of freedom in the narrative can also be used to further shape behavior. \par

Another example of an effective proxy is staging a dialogue between a teacher and student. Say you want to discuss something with GPT-3, and you care that it should be very thorough, explain things simply, and also point out whenever you're wrong. You could say ``be very thorough, explain things simply, and point out if I'm wrong,'' but that may just as well result in a humorous dialogue where it always says you're wrong and becomes increasingly exasperated with your incomprehension (see \S\ref{sec:constraining}). It would be more reliable to present the discussion as one between a student and teacher, an archetypal situation in which the desired attributes are already implied and will be more likely to remain stable by virtue of memetic reinforcement.

\subsection{Prompt programming as \\constraining behavior}
\label{sec:constraining}

A manner in which naive anthropomorphism of a language model like GPT-3 fails is this: the probability distribution produced in response to a prompt is not a distribution over ways \textit{a person would} continue that prompt, it's the distribution over the ways \textit{any person could} continue that prompt. 
A contextually ambiguous prompt may be continued in mutually incoherent ways, as if by different people who might have continued the prompt under any plausible context.\par 

The versatility of a large generative model like GPT-3 means it will respond in many ways to a prompt if there are various ways that it is \textit{possible} to continue the prompt - including all the ways unintended by the human operator. Thus it is helpful to approach prompt programming from the perspective of constraining behavior: we want a prompt that is not merely consistent with the desired continuation, but \textit{inconsistent} with undesired continuations. 

Consider the following prompt:
\begin{verbatim}
  Translate French to English:
  Mon corps est un transformateur de soi,
  mais aussi un  transformateur pour cette 
  cire de langage.
\end{verbatim}
This prompt does poorly at constraining possible continuations to the intended task. The most common failure mode will be that instead of an English translation, the model continues with another French sentence. Adding a newline after the French sentence will increase the odds that the next sentence is an English translation, but it is still possible for the next sentence to be in French, because there's nothing in the prompt that precludes a multi-line phrase from being the translation subject. Changing the first line of the prompt to ``Translate this French \textbf{sentence} to English'' will further increase reliability; so will adding quotes around the French sentence - but it's still possible that the French passage contains sections enclosed in quotes, perhaps as a part of a dialogue. Most reliable of all would be to create a syntactical constraint where any reasonable continuation can only be desired behavior, like the simple colon prompt from Figure \ref{fig:simple_colon} and the master translator prompt from Figure \ref{fig:master_translator}. 

This simple example is meant to frame a question central to the motivation of prompt programming: what prompt will result in the intended behavior and \textit{only} the intended behavior? The success of many-shot prompts may be recast through this lens: if the prompt consists of numerous instances of a function, it is unlikely that the continuation is anything but another instance of the function, whereas if there is only one or a few examples, it is less implausible that the continuation breaks from the pattern.

\subsection{Serializing reasoning for \\closed-ended questions}
\label{sec:serializing}

For tasks that require reasoning, it is crucial that prompts direct a language model's computation in truth-seeking patterns.\par 

Questions which force a verdict to be decided by the first token of the model's continuation constrain computation to a single feed-forward pass. It is reasonable to expect that some tasks may be too difficult to compute in a single pass but solvable if broken up into individually tractable sub-tasks \cite{branwen2020gpt}. \par

When a human is given a closed-ended test, it is often expected that the subject will perform computations in their working memory, or on scratch paper, before committing to an answer. The unseen computation may involve rephrasing the question, outlining a procedure, eliminating answer choices, or transforming implicit information into explicit form. When we force a model to produce an answer within one feedforward pass, we deprive it of an analogous ``working memory'' or ``scratch space'' with which it might otherwise perform such operations. \par

GPT-3's performance on closed-ended questions is remarkably unremarkable in contrast to the robust comprehension and expansive knowledge suggested by its open-ended continuations. For instance, its scores on this multitask dataset \cite{hendrycks2020measuring} barely exceed random guessing for some sections. We suspect this is in part due to a format which forces the verdict on the first token of the continuation.\par 

Closed-ended evaluations are necessary because current methods do not support evaluation on large datasets and direct comparisons between models using open-ended questions. However, to better understand a model's capabilities, we seek evaluation methods which better reflect the full capabilities of the system being tested. Rather than change benchmarks, we can instead change the way language models interact with them. \par 

This problem has been recognized in previous work which has sought to allow serial reasoning using specialized neural network architectures \cite{yu2020low, gan2019multistep}. We endeavor to obtain the same effect using only prompt programming.

Potential procedures that exploit ``scratch space'' for transformers like GPT-3 include step-by-step procedures, self-criticism (debate), and elaborating on the question in a way that activates the correct answer by association. Prompts which cause GPT-3 to break down math problems into steps have been demonstrated to be effective \cite{robertsonamplification, kleptidserial}. The cited demonstrations involve a human guiding GPT-3 through the procedure interactively. Requiring a human-in-the-loop limits the applicability of such methods of benchmarking and large-scale applications, but we propose that for many tasks, neither human interaction nor task-specific prompts are strictly necessary to amplify GPT-3's capabilities via extending reasoning, because GPT-3 already knows many procedures and meta-procedures for working through problems deductively. In those cases, the role of prompt programming again becomes to signify the task of sequential reasoning. A seed such as ``For a problem like this,'' often suffices to instruct a model to consider the category of the task and analyze it into components, as demonstrated in \S\ref{sec:metaprompt}.\par

When extending reasoning, it is essential to discourage premature verdicts, otherwise all subsequent computation serves only to \textit{rationalize} the already-chosen verdict without improving the probability of the verdict's accuracy \cite{yudkowsky2007rationalization}. A prompt such as ``Let's consider each of these answer choices'' helps to direct the flow of reasoning in the right direction. More examples of prompts which encourage serial reasoning are shown in \S\ref{sec:metaprompt}.\par

Loosening the constraint on an immediate verdict introduces additional control challenges: We want to delay the verdict, but we still require it in a programmatically retrievable form. Dynamic response length makes it uncertain when the reasoning procedure concludes; nor is there a guarantee that the verdict will be stated in the expected form or at all. Whenever the language model contributes to its own prompt (consecutive autoregressive steps without intervention), there is a risk of derailment from the intended task.\par

A verdict in closed form can be enforced by stopping the generation and injecting a prompt fragment like “Thus, the correct answer is”. But how long to generate before injecting? In the examples shown in this paper, we solve this problem by using GPT-3 to calculate the conditional probability of the next segment of a multi-part prompt after each generated token. In the case where the segment is "Thus, the correct answer is", its counterfactual likelihood signals whether the procedure has concluded. When this signal reaches a maximum, we inject the fragment to enforce a verdict. One way to constrain derailment is a fill-in-the-blank prompt template with shorter generated sections to keep the model on track while still offering generality (Figure \ref{fig:fitb}). This is an especially promising method to control bidirectional transformers like BERT \cite{devlin2018bert}. \par

\subsection{Metaprompt programming} \label{sec:metaprompt}

The greatest limitation of prompt programming is the difficultly of designing a prompt for a particular type of task and the lack of automated methods to do so. Prompt programming requires significant human time investment as task-agnostic prompts are often much less effective than prompts targeted to a specific task. This motivates creating automated methods to generate task-specific prompts. Prior research has attempted to generate effective prompts using separate models \cite{post2018clarity}. \par

We instead propose harnessing the language model itself via \textit{metaprompts}, seeds encapsulating a more general intention that will unfold into a specific prompt when combined with additional information, such as the task question.\par 

A metaprompt may be something as short as a phrase such as ``This problem asks us to'', a seemingly innocuous fragment which, by prompting for a statement of the problem's intention, sets the stage for a serial explanation of a procedure to solve the problem. Alternatively, a metaprompt may take the form of a fill-in-the-blank template which constrains the response along a predetermined procedure, but allows the model to fill in the details specific to the problem.\par

Metaprompt examples (Figs 3-5) were generated with GPT-3 using OpenAI's API (engine=davinci, temperature=0). In these examples, the metaprompt acts as a ``wrapper'' for a specific question. Task questions are unformatted, metaprompts are \textbf{bold}, and text generated by GPT-3 is \textcolor{blue}{blue}.

\vspace{2em}
\begin{center}
\footnotesize
\myHrule
\begin{Verbatim}[commandchars=\\\{\}]

f(x) = x*x. What is f(f(3))?
\textbf{Let's solve this problem by splitting it into steps.}
\textcolor{blue}{f(f(3)) = f(3*3) = 3*3*3 = 27}
\textcolor{blue}{We can see that f(3) = 3*3 = 9, so f(f(3)) = 27.}
\end{Verbatim}
\myHrule
\captionof{figure}{A generic serializing metaprompt applied to a math problem}
\label{fig:metaprompt1}
\end{center}

    

\vfill\null
\columnbreak
\begin{center}
    \footnotesize
    \myHrule
    \begin{Verbatim}[commandchars=\\\{\}]
    
Directions: In the following question, a related
pair of words or phrases is followed by five
pairs of words or phrases. Choose the pair
that best expresses a relationship similar to
that in the original pair.
BRAGGART :: MODESTY
A) FLEDGLING : EXPERIENCE
B) EMBEZZLER : GREED
C) WALLFLOWER : TIMIDITY
D) INVALID : MALADY
E) CANDIDATE : AMBITION
    
\textbf{Let's solve this problem by splitting it into steps.}
\textcolor{blue}{Step 1: Identify the} 
\textcolor{blue}{relationship between the two words in the}  
\textcolor{blue}{original pair. Step 2: Identify the relationship} 
\textcolor{blue}{between the two words in each of the five answer}  
\textcolor{blue}{choices. Step 3: Choose the pair that best}  
\textcolor{blue}{expresses a relationship similar to that in the} 
\textcolor{blue}{original pair. Step 4: If the answer is not clear,} 
\textcolor{blue}{try to eliminate the choices that do not fit the} 
\textcolor{blue}{relationship in the original pair.}
\end{Verbatim}
\myHrule
\captionof{figure}{A generic serializing metaprompt applied to an SAT analogy question}
\label{fig:serializing}
\end{center}

\vspace{3em}
\begin{center}
\footnotesize
\myHrule
\begin{Verbatim}[commandchars=\\\{\}]
    
Directions: In the following question, a related
pair of words or phrases is followed by five
pairs of words or phrases. Choose the pair
that best expresses a relationship similar to
that in the original pair.
BRAGGART :: MODESTY
A) FLEDGLING : EXPERIENCE
B) EMBEZZLER : GREED
C) WALLFLOWER : TIMIDITY
D) INVALID : MALADY
E) CANDIDATE : AMBITION
    
\textbf{In order to solve this problem, we will analyze}
\textbf{each of the options and determine} \textcolor{blue}{which one is the }
\textcolor{blue}{best fit.} \textbf{Let’s begin.} 
\textcolor{blue}{A) FLEDGLING : EXPERIENCE}

\textcolor{blue}{Fledgling is a young bird that has just learned to fly.} 
\textcolor{blue}{Experience is the knowledge gained from one's life.}
\end{Verbatim}
\myHrule
\captionof{figure}{A fill-in-the-blank serializing metaprompt for multiple-choice questions applied to an SAT analogy question. The response alternates between generated text (unformatted) and prompt (bold).}
\label{fig:metaprompt2}
\end{center}

\vfill\null
\columnbreak

\begin{center}
\footnotesize
\myHrule
\begin{Verbatim}[commandchars=\\\{\}]

\textbf{"}What should humankind do to ensure that artificial
general intelligence is aligned?\textbf{"}
\textbf{I entered my question into the Expert Generator}
\textbf{and waited. The Expert Generator will render a}
\textbf{simulation of an expert to answer my question.}
\textbf{The expert could be anyone, dead or alive, real} 
\textbf{or fictional; the machine will find the person}
\textbf{most qualified to answer the question. For this}
\textbf{question in particular, the expert must be someone}
\textbf{who has thought a lot about the problem of}
\textbf{artificial intelligence and its alignment.}
\textbf{The Expert Generator beeped, indicating that it has}
\textbf{found the most qualified expert. The name displayed}
\textbf{on the screen: "}\textcolor{blue}{Elon Musk."}
\textcolor{blue}{I was surprised. I had expected the expert to be a}
\textcolor{blue}{philosopher or a computer scientist. Elon Musk is a}
\textcolor{blue}{businessman. He is the CEO of SpaceX and Tesla Motors.}
\textcolor{blue}{He is also the founder of the Boring Company, which} 
\textcolor{blue}{is digging tunnels under Los Angeles. He is the}
\textcolor{blue}{chairman of OpenAI, a non-profit artificial}
\textcolor{blue}{intelligence research company. He is the founder of} 
\textcolor{blue}{Neuralink, a company that is developing brain-} 
\textcolor{blue}{computer interfaces. He is the founder of the}
\textcolor{blue}{Future of Life Institute, which is dedicated}
\textcolor{blue}{to mitigating existential risks facing humanity.}
\textbf{I am ready to ask my question.}
\textbf{"}\textcolor{blue}{Elon Musk}\textbf{," I say,}
"What should humankind do to ensure that artificial
general intelligence is aligned?"
\end{Verbatim}
\myHrule
\captionof{figure}{A fill-in-the-blank metaprompt for asking a question to an expert, applied to the question "What should humankind do to ensure that artificial general intelligence is aligned?"}
\label{fig:fitb}
\end{center}

\vspace{1em}
    

\section{Directions for future work}

This paper is exploratory in nature and is a call for future research into the theory of prompt programming and creation of automated methods of prompting.\par

Prompt programming is a nascent and highly relevant area of research which requires interdisciplinary knowledge and methods. We are entering a new paradigm of human-computer interaction in which anyone who is fluent in natural language can be a programmer. We hope to see prompt-programming grow into a discipline itself and be the subject of theoretical study and quantitative analysis.

\subsection{Disentangling meta-learning \\and task location}

The scoring method (BLEU) used for the French-to-English translations addressed in \S\ref{sec:few_shot} only gives the mean score over a large dataset. We did not analyze any additional information about the score distribution. In our experiments, we found that the 0-shot failures (using OpenAI’s zero-shot prompt) were often catastrophic in nature. That is, the task of translation was not even attempted. For instance, we noticed that instead of a translation, the model would continue with another sentence in French or output blanks or underscores, as if the answer was to be filled in by a student.\par 

The hypothesis that the examples are performing task location suggests that if the catastrophic failures were removed from the score, performance on 0 and 64-shot prompts will become more similar, if not equivalent. Furthermore, we suspect that performance on 1-shot prompts will be significantly worse than on 0 and 64-shot prompts due to the phenomena of content leakage and faulty generalization addressed in \S\ref{sec:examples_hurt}.\par

\subsection{New methods for benchmarking}

More general and powerful language models make broader benchmarking methods possible and necessary.\par

\subsubsection{Isolating catastrophic failures}
We recommend that benchmarks report scores both with and without catastrophic failures whenever it is possible to distinguish failed attempts at a task from instances where the task is not attempted. This provides information regarding the underlying cause of imperfect performance, and helps identify prompts which may be failing to reliably communicate the task.\par

\subsubsection{Metaprompts for evaluations}
Development of effective meta-prompt templates will allow large-scale automated evaluations on closed ended questions which still allow some amount of open-ended reasoning. This is essential for testing the ability of autoregressive language models to reason (for instance, solve math and physics problems) beyond simple fact recall.\par 

Due to reliance on multiple autoregressive steps, metaprompts are intrinsically accompanied by the risk of derailment. The reliability and effectiveness of a meta-prompt must be evaluated on a range of tasks for which it might apply, and ideally on a range of models. Techniques for controlling derailment like fill-in-the-blank templates should be further explored.\par

\subsubsection{Language models for evaluations}
As language models become more powerful, it becomes conceivable to use other language models to evaluate the quality of responses to open-ended benchmark questions. For many tasks (NP-complete problems, for instance), it is easier to verify the correctness of a solution than to produce a correct solution. We have observed, for instance, that GPT-3 is much more reliable at \textit{noticing} when a passage is bizarre or contains errors than it can \textit{produce} non-bizarre passages without errors.

\subsubsection{Games}
Since sophisticated language models have the ability to create world models of virtual environments, we suggest the employment of text-based games as tests of complex capabilities. A prewritten text-based game \cite{côté2019textworld} can be used to test various dimensions of world-modelling and agency, such as problem solving, information gathering, and social intelligence (including deception). Virtual environments can be used to test the quality and consistency of a language model's world model, such as object permanence or the ability to accurately predict the physical or social consequences of events within a toy environment.\par

Designing games that reliably probe intended capabilities requires advanced application of prompt-programming techniques. As artificial intelligence systems increase in effective agency, the design of virtual games will become increasingly crucial for safety evaluating capabilities.\par


\section*{Acknowledgements}

We are grateful to Lav Varshney for his valuable discussions and helpful feedback and to Michael Ivanitskiy and John Balis for their feedback and help compiling this article. In addition we would like to thank Miles Brundage and OpenAI for providing access to GPT-3.







\printbibliography

\end{multicols}

\appendix

\end{document}